\documentclass[letterpaper, 10 pt, journal, twoside]{ieeetran} 
\usepackage{times}

\usepackage{svg}

\usepackage{svg}
\usepackage{color}
\usepackage{transparent}
\usepackage{graphicx}
\usepackage{microtype}  
\usepackage{import}
\usepackage{wrapfig}
\usepackage{algorithm}
\usepackage{diagbox}
\usepackage{subcaption}
\usepackage{multicol}
\usepackage{tablefootnote}
\usepackage[normalem]{ulem} 
\usepackage[noend ]{algpseudocode} % algpseudocode is now commented out in ln17 of packages2include.text

\usepackage{enumitem} % tanamy added this for custom labels

\input{abkuerzung.sty}

\IEEEoverridecommandlockouts                              % This command is only needed if 
\usepackage[utf8]{inputenc}
\usepackage{leftidx}
\usepackage{bm}
\usepackage{amssymb,amsmath,amsfonts,empheq}
\usepackage{graphicx}
\usepackage{siunitx}
\usepackage{multicol}
\usepackage{algorithm}
\usepackage{type1cm}
\usepackage{pifont}% http://ctan.org/pkg/pifont
\usepackage{lettrine}
\usepackage{color}
\usepackage{cite}
\usepackage{breqn}
\usepackage{tikz}
\usepackage{array}
\usepackage{import}
\usepackage{amsmath}
\usepackage{amssymb}
\usepackage{bm} 
\usepackage{booktabs}
\usepackage{optidef}
\usepackage{graphicx}
\usepackage{xcolor}
\usepackage{siunitx}
\usepackage{fixltx2e}
\usepackage{multirow}
\usepackage{cite}
\usepackage{import}
\usepackage[draft]{hyperref}
\usepackage{subcaption}
\usepackage{microtype}
\usepackage{tabularx}
\usepackage{booktabs}
\usepackage{comment}
 \usepackage{microtype}
 \usepackage{xifthen}

\usepackage{multirow}
\usepackage{scalerel}
\usepackage{fixltx2e}

\usepackage{orcidlink}

\makeatletter%change heading spacings
\renewcommand{\section}{\@startsection{section}{1}{\z@}{0.6ex plus 0.6ex minus 0.3ex}%
	{0.4ex plus 0.6ex minus 0ex}{\normalfont\normalsize\centering\scshape}}%
\renewcommand{\subsection}{\@startsection{subsection}{2}{\z@}{0.6ex plus 0.6ex minus 0.5ex}%
	{0.3ex plus 0.5ex minus 0ex}{\normalfont\normalsize\itshape}}%
\makeatother

\renewcommand{\baselinestretch}{0.90}
\newcommand{\bigO}[1]{\mathcal{O}\left(#1\right)}

\newcommand{\fr}[2]{%\ensuremath{%
    \bm{f}%
    _{%
    \text{r}
        \if\relax\detokenize{#1}\relax
        \else
            ,#1
        \fi
    }%
    \if\relax\detokenize{#2}\relax
    \else
        ^{(#2)}
    \fi
}%}

\newcommand{\fa}[2]{%\ensuremath{%
    \bm{f}%
    _{%
    \text{a}
        \if\relax\detokenize{#1}\relax
        \else
            ,#1
        \fi
    }%
    \if\relax\detokenize{#2}\relax
    \else
        ^{(#2)}
    \fi
}%}

\newcommand{\pose}[3][]{%\ensuremath{%
  \bm{\xi}%
  _{%
    \if\relax\detokenize{#1}\relax
      \if\relax\detokenize{#2}\relax
      \else
        #2%
      \fi
    \else
      \text{#1}%
      \if\relax\detokenize{#2}\relax
      \else
        , #2%
      \fi
    \fi
  }%
  \if\relax\detokenize{#3}\relax
  \else
    ^{(#3)}%
  \fi
}%}

\newcommand{\F}[2]{\ensuremath{%
    \bm{f}%
    _{
        \if\relax\detokenize{#1}\relax
        \else
            #1
        \fi
    }%
    \if\relax\detokenize{#2}\relax
    \else
        ^{(#2)}
    \fi
}}

\newcommand{\oid}[1]{%
  \raisebox{0.3ex}{\orcidlink{#1}}%
}
\newcommand{\SWR}{\mathcal{S}_W}

\newcommand{\Ns}{N_{\text{S}}}

\newcommand{\No}{N_{O}}

\newcommand{\Na}{N_{a}}

\newcommand{\Nk}{N_{k}}

\newcommand{\Nsw}{N_{sw}}

\newcommand{\Pw}{\mathcal{P}_{W}}

\begin{document}

\title{G-MAPP: GPU-accelerated Multi-Agent Planning and Perception for Reactive Motion Generation}
%
% author list
% \author{Tanmay Bishnoi$^{1}$, Riddhiman Laha$^{2,6}$, Tobias Löw$^{3}$, Jose Alex Chandy$^{4}$,\\ Luis F.C. Figueredo$^{4,2}$, and Sami Haddadin$^{5}$%

\author{Tanmay Bishnoi\oid{0009-0003-7830-256X}, Riddhiman Laha\oid{0000-0001-8527-0445}, \textit{Graduate Student Member, IEEE}, Tobias Löw\oid{0000-0003-4001-2770}, Jose Alex Chandy\oid{0000-0002-5840-5727}, \\Luis F. C. Figueredo\oid{0000-0002-0759-3000}, \textit{Senior Member, IEEE}, and Sami Haddadin\oid{0000-0001-7696-4955}, \textit{Fellow, IEEE}

%
% % manuscript info
% \thanks{Manuscript received: November, 13, 2025; Revised: February, 7, 2026; Accepted: March, 3, 2026.}%
% \thanks{This paper was recommended for publication by Editor Aniket Bera upon evaluation of the Associate Editor and Reviewers' comments.}%
% manuscript info
\thanks{Manuscript received: November, 13, 2025; Revised: February, 7, 2026; Accepted: March, 3, 2026. This paper was recommended for publication by Editor Aniket Bera upon evaluation of the Associate Editor and Reviewers' comments. \textit{(Corresponding author: Tanmay Bishnoi, email: tbishnoi@torontomu.ca.)}}%
%
% affil
\thanks{Tanmay Bishnoi is with the Department of Electrical, Computer, and Biomedical Engineering, Toronto Metropolitan University, Toronto, Canada. Riddhiman Laha is with the Munich Institute of Robotics and Machine Intelligence (MIRMI), Technical University of Munich (TUM), Munich, Germany, and also with the Institute for Experiential Robotics, Northeastern University, Boston, USA. Tobias Löw is with Idiap Research Institute, Martigny, Switzerland, and also with EPFL, Lausanne, Switzerland. Jose Alex Chandy and Luis F.C. Figueredo are with the CHART Group at the School of Computer Science, University of Nottingham, Nottingham, U.K.. Luis F. C. Figueredo is also with Munich Institute of Robotics and Machine Intelligence (MIRMI), Technical University of Munich (TUM), Munich, Germany. Sami Haddadin is with the Mohamed bin Zayed University of Artificial Intelligence (MBZUAI), Abu Dhabi, UAE.
}%
\thanks{Digital Object Identifier (DOI): see top of this page.}
}
%
% headers
\markboth{IEEE Robotics and Automation Letters. Preprint Version. Accepted March, 2026}{Bishnoi \MakeLowercase{\textit{et al.}}: G-MAPP: GPU-accelerated Multi-Agent Planning and Perception for Reactive Motion Generation}

\maketitle
\begin{abstract} Reactive motion generation in unstructured environments remains an open challenge in robotics. Due to the computational complexity of collision-free motion generation, existing methods either generate global trajectories for static scenarios, or employ models that make conservative assumptions about the environment. This paper identifies the primary bottleneck as the runtime performance demand of planning on high-fidelity environments, and the temporal integration between the perception and planning modules. Therefore, we propose a framework that does not compromise on runtime performance and world representations for perception and planning by accelerating world modeling and vector-field based planning using the GPU. This allows us to achieve faster parallel state exploration for quasi-global trajectory planning, and tighter coupling of the perception-action loop in real-time for dynamic cluttered environments with off-the-shelf depth sensors. We quantitatively evaluate the computation-time and success rate differences for the CPU and GPU versions of our planner, and perform qualitative evaluations of our coupled framework using real-world experiments on a $7$-DoF Franka Emika robot. Experimental results demonstrate that our GPU-based framework achieves up to a $5$x speedup over the CPU version and successfully avoids collisions across both trivial and challenging physical world scenarios. The implementation is available at: https://github.com/chart-research/g-mapp
\end{abstract}


\IEEEpeerreviewmaketitle

%%%%%%%%%%%%%%%%%%%%%%%%%%%%%%
% INTRODUCTION
%%%%%%%%%%%%%%%%%%%%%%%%%%%%%%
% -- Introduction ---
\section{Introduction}

% -- Intro --
\IEEEPARstart{R}{obotic} systems entering collaborative spaces need capabilities to tackle uncertainties that are inherent to these environments~\cite{koditschek2021robotics}. It is an established belief that designing capable online systems that can handle dynamic unstructured environments require a tight coupling between perception and planning for motion generation \cite{brooks1990elephants,eppner2016lessons,kappler2018reactive}.   
However, perception and planning come with their own set of practical demands, which hamper efficient real-time integration. These different requirements lead to the traditional sense-plan-act paradigm, which, although successful for static scenes, lacks the reactivity features needed for human-centered environments~\cite{gat1998three}. In this work, we propose a novel reactive planning architecture that effectively couples perception, planning, and action under a real-time framework. Our strategy promotes task-space exploration using a GPU-parallelized vector-field through a \emph{geometry-aware parameterization} for robot kinematics. More specifically, rigid body motions and reactive forces are encoded using geometric algebra (GA), which naturally encodes the full $6$D task space, thus bridging the gap between geometric representation and reactive planning in the operational space.

% --- SBMP Review ---
% To address the NP-hard\footnote{Polynomial-time infeasibility for guaranteed collision-free convergence.} nature of motion planning~\cite{reif1994motion}, researchers have focused on probabilistic completeness, i.e., ensuring convergence guarantees. 
%
Exact motion planning is NP-hard and typically infeasible to solve in polynomial time ~\cite{reif1994motion}. Researchers have, thus, focused on probabilistic completeness, i.e., ensuring convergence guarantees. Sampling-based motion planners (SBMPs), such as the rapidly-exploring random trees (RRT)\cite{lavalle1998RRT}, probabilistic roadmaps (PRM)\cite{kavraki1996PRM}, and their variants\cite{Karaman2011RRTStar,Kuffner2000RRTConnect}, have proven to be powerful and versatile tools~\cite{orthey2023sampling}. They make high–DOF motion planning tractable by trading strict completeness for probabilistic guarantees, allowing for broad coverage of the environment with the trade-off of time-performance. SBMPs employ quasi-random sampling strategies for state-space exploration and graph construction. However, they often exhibit asymptotically poor runtime performance in \emph{narrow passage} and geometry-aware environmental constraint problems, where critical regions of the state space (necessary for connectivity) are rarely sampled~\cite{livnat2024tight}. Recent work on SBMPs focuses on accelerating algorithmic ``primitives’’ used within these algorithms \cite{thomason2024vamp, ramsey2024capt, huang2025prrtc} to abate the performance impact of complex motion planning situations.
\begin{figure}[t]
    \centering
    \includegraphics[width=\columnwidth]{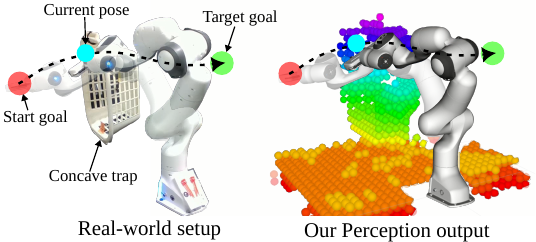}
    \caption{A snapshot of our real-time perception and planning pipeline. The environment is captured through a GPU-accelerated vision system that fuses multi-camera inputs into a simplified, structured representation using geometric primitives (e.g., spheres). This low-latency abstraction enables the robot to perform continuous, reactive motion planning in cluttered and fast-changing environments.}
    \label{fig:front_page_graphic}
\end{figure}

% --- Reactive Review ---
In contrast, another line of research prioritizes safety and collision avoidance in dynamic scenarios. Traditional locally reactive or dynamical methods, often inspired by physical analogies, trade convergence guarantees for real-time performance gains. This makes them desirable for instantaneous reactions to dynamic environments~\cite{ngo2024joint}. Their simple, closed-form formulation enables direct evaluation of control laws with low computational overhead and deterministic outcomes, distinguishing them from the sense-plan-act architecture typical of most SBMPs. Among these methods, notably, are artificial potential fields (APF) \cite{khatib1986apf} and circular fields (CF) \cite{singh1996cf}. Most dynamical systems, however, suffer from the \emph{local minima} problem where the robot’s configuration as a point in a potential field becomes trapped in a local minima and fails goal convergence~\cite{panagou2014motion}. Recent research work in the field focuses on extending locally reactive methods using novel heuristic formulations and system design architectures to achieve global trajectory planning \cite{becker2021circular,laha2023pmaf}. Notably, the \emph{multi-agent} strategy (for operational space exploration), which closes the gap between reactivity and goal convergence.

% --- Perception Review ---

Despite these advantages, most reactive methods still rely on perception pipelines that are slow and decoupled~\cite{laha2023pmaf}. In these cases, computing distances to numerous obstacles or dense point clouds--essential for generating virtual forces--becomes exponentially more demanding as scene complexity increases~\cite{vasilopoulos2023ramp}. Consequently, reactive planning methods face a trade-off: they must either restrict exploration or sacrifice real-time performance, often simplifying distance-based force field computations through overly conservative assumptions about the number and shape of obstacles. Moreover, planning methods designed and validated \emph{in abstracto} can fail in practice without careful consideration of the perception scheme or its coupling with the planning component \cite{eppner2016lessons, bohg2017interactive}. Some systems (at their core) employ implicit world models \cite{yang2025deep,you2025inspection}, resembling a strict sensorimotor coupling paradigm in which raw sensor values are mapped to motor commands. Others explicitly construct models of the environment, utilizing prior knowledge of the world’s geometry \cite{eckhoff2023perception, gong2025geopf} or its dynamics \cite{ai2025dynamics} to achieve efficient reactive control. At a high-level, optimal perception module design remains an open-ended challenge \cite{eppner2016lessons}. However, certain requirements are essential, such as a low end-to-end latency of operation, tight system integration \cite{bohg2016interlocking}, and adequate representational fidelity, either free from or compensating of sensing-related artifacts. Extensibility to multi-sensor integration can enhance embodiment capabilities but increases computational demands.

% In the existing literature, there remains a lack of a reactive planner that is capable of interfacing high-fidelity representations across diverse control laws and meets end-to-end computational efficiency and robust performance requirements. This letter attempts to bridge that gap by demonstrating that our GPU-accelerated perception-planning framework satisfies online real-world manipulation demands in dynamic unstructured environments, and achieves a balance between modularity and tight system integration. This differentiates our work from prior work in the area \cite{kaldestad2014collision}. Our main contributions are:
% \begin{enumerate}[label=C\arabic*.]
%     \item A low-latency GPU-accelerated perception module that fuses multi-camera point clouds into a structured representation, allowing real-time capture of fast-changing physical worlds.
%     \item A GPU-parallelized multi-agent task-space planner that evaluates and selects motion strategies in cluttered, high-fidelity environments. Our method, in contrast to classical reactive approaches, emphasizes predictive strategy selection. This is crucial for maintaining responsive motion generation under uncertainty in fast-changing environments.
%     % \item A GA-based formulation for vector-field planning that naturally respects the manifold structure of rigid-body motion. Using GA allows us to propagate a coupled translation and orientation representation, and affords us the use of geometric motion primitives common to real-world robotic problems.
% \end{enumerate}

In the existing literature, there remains a lack of a reactive planner that is capable of interfacing high-fidelity representations across diverse control laws and meets end-to-end computational efficiency and robust performance requirements. This letter attempts to bridge that gap by demonstrating that our GPU-accelerated perception-planning framework satisfies online real-world manipulation demands in dynamic unstructured environments, and achieves a balance between modularity and tight system integration. This differentiates our work from prior work in the area \cite{kaldestad2014collision}.

Our contributions are twofold: i) we present a low-latency GPU-accelerated perception module allows real-time capture of fast-changing physical worlds into a structured representation with off-the-shelf depth sensors.
ii) we propose a \emph{GPU-parallelized multi-agent task-space planner} that can evaluate and select motion strategies in dynamic, cluttered, and high-fidelity environments.

% \begin{enumerate}[label=C\arabic*.]
%     \item A low-latency GPU-accelerated perception module that fuses multi-camera point clouds into a structured representation, allowing real-time capture of fast-changing physical worlds.
%     \item A GPU-parallelized multi-agent task-space planner that evaluates and selects motion strategies in cluttered, high-fidelity environments. Our method, in contrast to classical reactive approaches, emphasizes predictive strategy selection. This is crucial for maintaining responsive motion generation under uncertainty in fast-changing environments.
%     % \item A GA-based formulation for vector-field planning that naturally respects the manifold structure of rigid-body motion. Using GA allows us to propagate a coupled translation and orientation representation, and affords us the use of geometric motion primitives common to real-world robotic problems.
% \end{enumerate}

\section{Mathematical Problem Statement} \label{sec:problem_statement}

Let the robot operate in a task-space $\mathcal{T} \subset \text{Spin}$(3)$ \ltimes \mathbb{R}^3$, then $\pose[s]{}{},\pose[g]{}{} \in \mathcal{T}$ represent the start and goal end-effector poses, respectively. We consider a real-time motion planning scenario in an unstructured, previously unknown, dynamically evolving environment. The broad objective is to find a trajectory $\mathbb{T} = (\pose[]{}{1}, \ldots, \pose[]{}{k}, \ldots, \pose[]{}{N})$, 
where $k, N \in \mathbb{Z}^{+}$ and $\pose[]{}{k} \in \mathcal{T}^{\text{free}}$, 
with $\mathcal{T}^{\text{free}} \subset \mathcal{T}$ denoting the collision-free subset.

In this letter, we would like to focus on the following three distinct sub-problems that we consider essential for real-world reactive motion generation:

\textbf{Problem P}$\mathbf{1}$ (Perception Constraints). Perception for reactive motion planning has a set of associated challenges that a pipeline should address to be considered suitable. These are: i) modularity (must pair with other reactive planners and additional sensors), ii) occlusion-free environment sensing, iii) high-fidelity representation, iv) low spatio-temporal artifacts and noise, and v) low-latency processing time. 

\textbf{Problem P}$\mathbf{2}$ (Control Law Time-costs). A reactive planner must be able to compute virtual forces calculated from the high-fidelity representations generated by the perception pipeline under a \emph{time budget} to maintain reactive control of the robot. We consider an appropriate frequency of real-time robot control to be $100$ Hz  \cite{laha2023pmaf}, which the planner must respect---irrespective of scene complexity. %This implies that an ideal planner needs to compute the forces under a time budget with a time complexity of $\approx O(1)$ with respect to the number of obstacles---i.e., the planning time should be quasi-constant (independent), despite the environment complexity to prevent loss of performance.

\textbf{Problem P}$\mathbf{3}$ (Parallel Task-space exploration). To account for the local minima problem, measures should be taken to ensure the task-space is actively explored and alternate trajectories are considered in parallel to the current real-time robot control strategy. The task-space exploration, similar to Problem P$2$, must be done under a time budget, and (ideally) should not be affected by the complexity of the scene or the number of parallel explorations.

We can now articulate the discussion above as follows.\\
\noindent \textbf{Problem Statement.} \textit{Given the start and goal poses $\pose[s]{}{},\pose[g]{}{}\in\mathcal{T}^{free}\subset \mathrm{Spin}(3) \ltimes \mathbb{R}^3$, a number of (exteroceptive) sensors $\Ns$, and a number of parallel ``task-space explorers'' $\Na$. Determine a suitable representation for motion planning $\SWR$, control forces, and prospective explored task-space paths, in real-time for reactive robot control.}

% \todo{add background section}
%%%%%%%%%%%%%%%%%%%%%%%%%%%%%%
% PRELIMS 
%%%%%%%%%%%%%%%%%%%%%%%%%%%%%%
\section{ Background and Preliminaries}\label{sec:mathematical_background}
\begin{figure*}[!h]
    \centering
    \includegraphics[width=\textwidth]{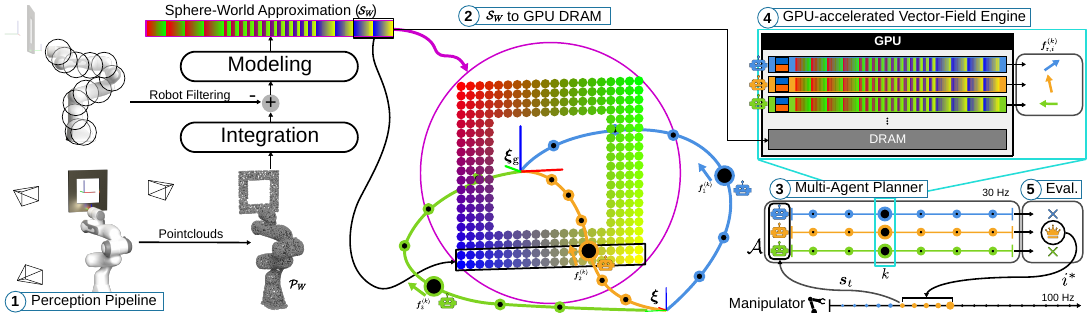}
    \caption{Overview of our coupled real-time GPU-accelerated perception and planning framework. The perception pipeline \textcircled{1} performs transformation of multiple synchronous point cloud frames into $\SWR$ array that is loaded onto GPU DRAM \textcircled{2}. At the start of a planning cycle, the multi-agent planner \textcircled{3} polls real-time state data ($\bm{s}_t$) on the robot and sets the multi-agents. It queries the GPU for the repulsive force vector, $\fr{i}{}$, to calculate the end effector force ($\F{i}{}$) for each multi-agent at each step of the planning cycle \textcircled{4}. At the end of the planning cycle, the agent evaluator selects the best agent,  $i^{*}$, that is used by the manipulator in real-time \textcircled{5}.}
    \label{fig:complete_system_overview}
\end{figure*}

\begin{figure}[htbp]
    % \centering
    \includegraphics[width=\columnwidth]{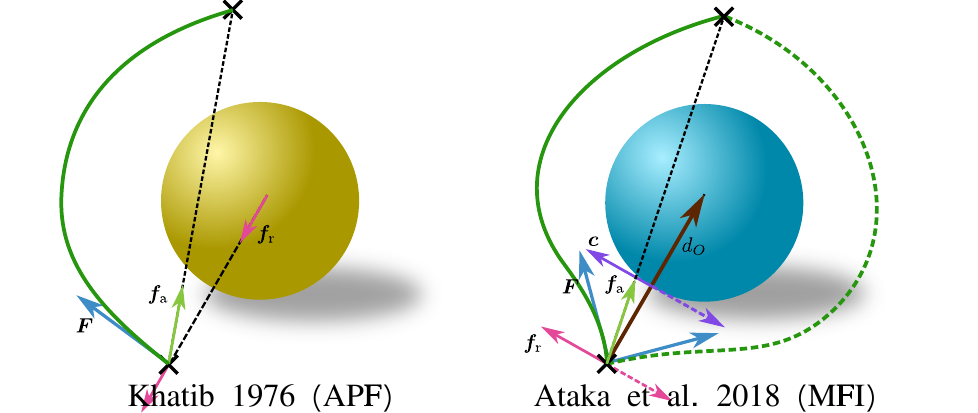}
    \caption{Schematic illustration depicting the essential vector formulations for two widely-used vector field heuristics for determining an avoidance direction. For each heuristic, the violet vector represents the repulsion force, and the dark red indicates the net force.}
    \label{fig:vector_visualizations_apf_and_mfi}
\end{figure}

The reactive methods, inspired by electromagnetic fields, can allow rotation around the obstacles (like a swirl) without affecting the total energy of the system\cite{ataka2018mfi, ataka2022mfi, laha2021reactive}. Still, as pointed out in~\cite{becker2021circular}, the traditional velocity (vector) fields, which are usually designed for real Euclidean space, suffer from two main drawbacks: (a) local minima artifacts, and (b) limit cycles (the robot trajectory spirals into itself \emph{ad infinitum}). Thus, in certain cases, the robot does not reach the goal. 

Therefore, we adopt a multi-agent approach \cite{becker2023informed,laha2023pmaf} where we spawn multiple agents during a planning cycle, each with a different heuristic (control law) and/or parameter set, to collectively explore a partially known environment in parallel. Each virtual agent samples a distinct trajectory in the task-space based on its programming. These agents span a diverse space of behaviors, probing multiple regions of the state space and exploring parts of the environment. From a distributional coverage point of view, they act like a set of dynamic modes. In the following subsection, we define the heuristics which these virtual agents employ.

\subsection{Vector Field Heuristics} \label{sec:vector_field_heuristics}

% \paragraph{Heuristic 1}

\textbf{Heuristic} $\mathbf{1}$ (APF). We incorporate an APF-based heuristic, offering a classical yet computationally efficient reactive behavior.
In this formulation, obstacle avoidance is handled via a repulsive field, active within a threshold distance $r_d$. Given the distance of the agent to the closest point on the obstacle surface $d_O$, the repulsive force is defined as
\begin{equation}
    \label{eq:apf}
    \fr{}{} =
    \begin{cases}
        k_f \left( \frac{1}{\| \bm{d} \|} - \frac{1}{r_d} \right) \frac{1}{\| \bm{d} \|^3} \, \bm{d}, & \text{if } d_O < r_d, \\
        \bm{0}, & \text{otherwise},
    \end{cases}
\end{equation}
where \( \bm{d} = \bm{p} - \bm{p}_O\) represents the distance between the agent and obstacle centers, and \(k_f > 0\) is the repulsion gain.

\textbf{Heuristic} $\mathbf{2}$ (MFI). In the CF approach, the robot is modeled as a charged particle moving in artificial electromagnetic fields and is subject to the following modified version of the Lorentz force where $r_O$ is radius of spherical primitive~\cite{laha2021reactive} 
\begin{equation}
    \label{eq:mfi}
    \fr{}{} = 
    \frac{k_f}{\lVert \bm{d} \rVert - r_O}\,
    \frac{\dot{\bm{d}}}{\lVert \dot{\bm{d}} \rVert}
    \times (\bm{c} \times \dot{\bm{d}})
\end{equation}
% where $k_{f}>0$ is a constant gain and $\bm{d} = \bm{p}_o - \bm{p}_a$ denotes the distance vector from the robot control point to the closest point on the obstacle surface $\bm{p}_o$, and the $\bm{c} \times \dot{\bm{d}}$ term denotes the magnetic field $B$ with artificial current $\bm{c}$. The artificial current defines the direction of the magnetic field, and thus, the direction of the CF force, that is, it defines the direction in which an obstacle is avoided. Conceptually, this can be thought of as a special case of a stream function that avoids obstacles via swirling. The current heading direction of the EE is taken into account when this heuristic comes into play. Here, the current EE (linear) velocity is projected onto the obstacle surface as elucidated 

where $k_{f}>0$ is a constant gain, $\bm{d} = \bm{p}_o - \bm{p}_a$ is the distance vector from the robot control point to the closest point on the obstacle surface $\bm{p}_o$, $r_{O}$ is the radius of the spherical primitive, and the $\bm{c} \times \dot{\bm{d}}$ term denotes the magnetic field $B$ with artificial current $\bm{c}$. The artificial current defines the direction of the magnetic field, and thus, the direction of the CF force, that is, it defines the direction in which an obstacle is avoided. The drawback of the CF method is that it fails for non-convex geometries and dynamics situations, as documented in \cite{ataka2018mfi}. We, therefore, implement the Magnetic-field inspired (MFI) heuristic with the parameterized current vector. % Conceptually, this can be thought of as a special case of a stream function that avoids obstacles via swirling. 
The current heading direction of the EE is taken into account when this heuristic comes into play. Here, the current EE (linear) velocity is projected onto the obstacle surface as elucidated in~\cite{ataka2018mfi}
\begin{equation}
    \bm{c} = \dot{\bm{d}} - \frac{\bm{d}_{O}}{{\|\bm{d}_{O}}\|}\left(\frac{\bm{d}_{O}}{\|\bm{d}_{O}\|} \cdot \dot{\bm{d}}\right).
\end{equation} 
Here, $\dot{\bm{d}}$ is the relative velocity between TCP and obstacle.

This artificial current for calculating the avoidance forces needs to be orthogonal to the obstacle surface normal.

\textbf{Other Heuristics}. The repulsive heuristics employed in our framework are inspired by [4]. While our approach is not restricted to these specific heuristic forms, they provide a principled and effective proof-of-concept. For the experiments reported in Table 1, we instantiate two agents per heuristic type, each with either a large or small detection radius. This design emulates a combination of broad-phase (collision-avoidance–focused) and narrow-phase (locally greedy) behaviors. As a result, multi-agent configurations consist of a heterogeneous ensemble
with varied heuristics.

% \cob{}

%%%%%%%%%%%%%%%%%%%%%%%%%%%%%%
% PERCEPT
%%%%%%%%%%%%%%%%%%%%%%%%%%%%%%
%%%%%%%%%%%%%%%%%%%%%%%%%%%%%%
\section{The Perception-Planning Framework} \label{sec:approach}
%%%%%%%%%%%%%%%%%%%%%%%%%%%%%%

\subsection{Approach Overview}

% --- Perception ---
Let the observations of the world be represented as
\begin{equation}
    \Pw = \bigcup_{i=1}^{N_{\text{S}}} \mathcal{P}_i ,
\end{equation}
where $\Ns$ is the number of unique sensors contributing to the world model, and $\mathcal{P}_i \in \mathbb{R}^{n_i \times 3}$ represents the point cloud captured by the $i$-th sensor, with $n_i \in Z^{+}$.
To support fast, geometry-aware planning in this space, we define a low-latency perception pipeline that transforms the fused point cloud $\mathcal{P}_W$  into a compact geometric abstraction
\begin{equation}
    \SWR = \left\{ \bm{o}_j = (\bm{p}_{O_{j}}, r_{O}) \in \mathbb{R}^3 \times \mathbb{R}_{>0} \;\middle|\; j = 1, \dots, \No \right\}
\end{equation}
where each $\bm{o}_j$ is a spherical primitive centered at $\bm{p}_{O_j}\in\mathbb{R}^{3}$ with radius $r_{O} \in \mathbb{R}_{>0}$ representing occupied space, and $\No\in\mathbb{N}$ being the number of such spherical primitives. This world model, enables the use of analytic potential-based vector fields directly. Note that $\SWR$ represents the occupied task-space, i.e. $\SWR = \mathcal{T}^{occupied}$. 

% --- Agent Dynamics ---
% We formulate the net wrench \footnote{Note: From this point onwards, we refer to wrenches as ($6$D) forces unless stated otherwise. The $\fr{}{}$ wrench has no rotational components; therefore, $\fr{}{}$ represents pure linear repulsive forces.} (forces and moments) on the agent $i$ as

We formulate the net wrench (forces and moments) with pure linear repulsive forces on agent $i$ as

\begin{equation}
\F{i}{} = \fa{}{}\Big(\pose[]{i}{k}\Big) + \fr{}{}\Big(\pose[]{i}{k}; \SWR\Big), \quad \pose[]{i}{k} \in \mathcal{T},
\end{equation}
where $\pose[]{i}{k}$ represents the pose of the agent at timestep $k$ in which the agent's motion is guided by:
\begin{itemize}
    \item a $C^{1}$ differentiable attractive potential $\fa{}{}\Big(\pose[]{i}{k}\Big)$ , pulling toward the goal, and
    \item a $C^{1}$ differentiable repulsive potential $\fr{}{}\Big(\pose[]{i}{k};\SWR\Big)$, defined over $\SWR$.
\end{itemize}
Since the attractive and repulsive potential functions are $C^{1}$ functions, their sum, the net force $\F{i}{}$, remains continuously differentiable as a $C^1$ function. This guarantees that the induced trajectories in the task-space are smooth in both position and velocity. 

The attractor force $\fa{}{}\Big(\pose[]{i}{k}\Big)\in \mathrm{Spin}(3)$ can be calculated as 
\begin{equation}
   \fa{}{}\Big(\pose[]{i}{k}\Big) = \bm{K}\bm{e}\Big(\pose[]{i}{k}\Big) - \bm{D}\bm{v}_i^{(k)},
\end{equation}
where $\bm{K}\in\mathbb{R}^{6\times 6}$ and $\bm{D}\in\mathbb{R}^{6\times 6}$ are the stiffness and damping gains, respectively. $\bm{v}_i^{(k)} \in \mathrm{Spin}(3)$ is the twist, i.e. the task-space velocity of the $i$-th agent at the current timestep $k$. The task-space error $\bm{e}$ is computed as the difference to the goal using the logarithmic map
\begin{equation}
    \bm{e}\Big(\pose[]{i}{k}\Big) = \log \left( \Big(\pose[]{i}{k}\Big)^{-1}\pose[g]{}{} \right).
\end{equation}

The $\bm{f}_r$ on the agent, is calculated using heuristics mentioned in Sec. \ref{sec:vector_field_heuristics} that define the interaction law between $\SWR$ and the agent.

We obtain a task-space acceleration $\dot{\bm{v}}_i^{(k)}$ from the net force on the $i$-th agent $\F{i}{}$ using a tunable inertia tensor $\bm{\mathcal{I}}\in\mathbb{R}^{6\times 6}$
\begin{equation}
    \dot{\bm{v}}_i^{(k)} = \bm{\mathcal{I}} \F{i}{}.
\end{equation}
Using a small timestep $\delta t$, we obtain the task-space velocity as
\begin{equation}
    \bm{v}_i^{(k)} = \bm{v}_i^{(k-1)} + \frac{1}{2} \delta_t \dot{\bm{v}}_i^{(k)}
\end{equation}
For a pointmass agent the new pose can be found as 
\begin{equation}
\pose[]{i}{k+1} = \pose[]{i}{k}\exp \Big(\delta_t \bm{v}_i^{k}\Big).
\end{equation}
For a robotic arm, the state ($\bm{s}_t$) additionally includes the current joint position $\bm{q}_i^{(k)}\in \mathbb{R}^{7}$ and velocity $\dot{\bm{q}}_i^{(k)}\in \mathbb{R}^{7}$. To update the state, we first compute the corresponding joint velocity as
\begin{equation}
    \dot{\bm{q}}_i^{(k)} = \bm{J}^{\dagger}\Big(\bm{q}_{i}^{(k)}\Big) \bm{v}_i^{(k)},
\end{equation}
where $\bm{J}^{\dagger}\Big(\bm{q}_{i}^{(k)}\Big)\in \mathbb{R}^{7\times 6}$ is the pseudo-Jacobian of the robotic arm at the current joint position. The new joint position is found as 
\begin{equation}
    \bm{q}_{i}^{(k+1)} = \bm{q}_{i}^{(k)} + \delta t \dot{\bm{q}}_{i}^{(k)}.
\end{equation}
Finally, the new pose is calculated using the forward kinematics (FK) function $f: \mathbb{R}^7 \to \mathrm{Spin}(3)\ltimes \mathbb{R}^3$ as
\begin{equation}
    \pose[]{i}{k+1} = f\Big(\bm{q}_i^{(k+1)}\Big).
\end{equation}
For parallel exploration of the task-space, we define a multi-agent planning scheme that generates candidate trajectories to evaluate using a set of planning agents $\mathcal{A} = \{ A_1, \dots, A_{\Na} \}$. The candidate trajectory of agent $A_i$ is
\begin{equation}
\mathbb{T}_i = (\pose[]{i}{1}, \dots, \pose[]{i}{\Nk}) \in \mathcal{T}, \quad \forall i \in \{1, \dots, \Na\},
\end{equation}
where $\pose[]{i}{k} \in \mathcal{T}$ denotes the $k$-th discretized pose along agent $A_i$'s trajectory $\mathbb{T}_{i}$, and $\Na,\Nk\in\mathbb{N}$ reflect the number of planning agents, and the number of planning steps in the scheme, respectively. 
Here, $\mathbb{T}_i$ is sampled from the set $\mathfrak{T}$, i.e. $\mathbb{T}_i \in \mathfrak{T}$ where $\mathfrak{T}$ is set of all trajectories. The candidate trajectories are evaluated against a cost function, $C: \mathfrak{T} \to \mathbb{R}$, that is subject to system and environmental constraints to select a parameter set for generating a smooth, short, and collision-free trajectory from $\pose[s]{}{}$ to $\pose[g]{}{}$.
% \todo{in general this is the correct idea, but $\mathbb{T}_{i}$ is only a particular trajectory from the set of all possible trajectories (needs to be given a symbol), then this function can be defined from set to set, and then called as $C(\mathbb{T}_{i})$ }\textcolor{green}{This is right, we need to change it. The function definition should be mapping the whole domain and not the instantiation. }

% --- Cost Function ---
\textbf{Cost Functions.} We evaluate each agent after $\Nk$ steps into the future and select the best agent according to a user-defined (task-dependent) cost function. We use a weighted combination of goal distance, obstacle distance, path length, and trajectory smoothness costs. This evaluation function can be readily extended to accommodate new scenarios or criteria.

Formally, the total cost for an arbitrary agent $i$ is 
\begin{equation}
\label{eq:cost_function}
C_i = \sum_{j=1}^4 w_j C_{i,j},
\end{equation}
where \(w_j\) are tunable scalars and \(C_{i,j}\) are defined as:
\[
\begin{aligned}
C_{i,1} &\equiv C_{\mathrm{goal}} &: &\ \text{Goal Distance Cost}, \\
C_{i,2} &\equiv C_{\mathrm{obs}} &: &\ \text{Obstacle Distance Cost}, \\
C_{i,3} &\equiv C_{\mathrm{path}} &: &\ \text{Path Length Cost}, \\
C_{i,4} &\equiv C_{\mathrm{smooth}} &: &\ \text{Trajectory Smoothness Cost}.
\end{aligned}
\]
The following subsections provide a detailed description of each cost component and its role in the planning objective.

\subsubsection{Goal Distance Cost} Penalizes distance from goal within detection radius $r_d$.
\begin{equation}
\label{eq:cost_goal}
    C_{\mathrm{goal}} =
    \frac{1}{\Nk} 
    \sum_{k=1}^{\Nk}
    \frac{\min(d_{\min}^{(k)}, r_d)}{r_d}
\end{equation}
where $d_{\min}^{(k)}$ is the Euclidean distance to the goal.

\subsubsection{Obstacle Distance Cost} \label{obs_dist_cost} Penalizes proximity to the closest obstacle within detection radius $r_d$.
\begin{equation}
\label{eq:cost_obs}
    C_{\text{obs}} = 
    \frac{1}{\Nk} 
    \sum_{k=1}^{\Nk}
    1-\frac{\min(d_{\min}^{(k)}, r_d)}{r_d}
\end{equation}
where $d_{\min}^{(k)}$ is the Euclidean distance to the closest obstacle.

\subsubsection{Path Length Cost}
Penalizes deviations of the total path length, $\mathcal{L}_{\text{path}} = \sum_{k=2}^{\Nk} \left\| \bm{p}^{(k)} - \bm{p}^{(k-1)} \right\|$, from the starting Euclidean distance to the goal, $d_{\min}^{(1)}$. 
\begin{align}
\label{eq:cost_path}
    C_{\mathrm{path}} &= 
    \begin{cases}
        1 - \frac{d_{\min}^{(1)}}{\mathcal{L}_{\mathrm{path}}}, & \mathcal{L}_{\mathrm{path}} \ge d_{\min}^{(1)} \\
        1 - \frac{\mathcal{L}_{\mathrm{path}}}{d_{\min}^{(1)}}, & d_{\min}^{(1)} > \mathcal{L}_{\mathrm{path}}
    \end{cases}
\end{align}

\subsubsection{Trajectory Smoothness Cost} Penalizes aggregate jerks, $\mathcal{J} = \sum_{k=4}^{\Nk} \left\| \bm{p}^{(k)} - 3\bm{p}^{(k-1)} + 3\bm{p}^{(k-2)} - \bm{p}^{(k-3)} \right\|^2$, along the trajectory.
\begin{gather}
\label{eq:cost_smooth}
    C_{\mathrm{smooth}} = \min\left(1, \frac{\mathcal{J}}{\Nk \Delta t^4 a_{\max}^2}\right)
\end{gather}
where the cost is normalized with the maximum possible acceleration $a_{\max}$ over timestep $\Delta t$.

% \textbf{Selection Mechanism.} Upon the computation of the total costs, the best agent is selected as
% %
% \begin{equation}
% \label{eq:best_agent}
%     i^\star = \arg\min_{i} \left\{ C_i \ :\  C_i < \alpha\, C_{\mathrm{best}} \right\},
% \end{equation}
% where $\alpha\in(0,1)$ is a user-defined scalar that tightens the selection criterion by ensuring that an agent is chosen as the new best agent only if its cost is sufficiently lower than $C_{\mathrm{best}}$. 
\textbf{Selection Mechanism.} Upon the computation of the total costs, the best agent is selected as $i^\star = \arg\min_{i} \left\{ C_i \ :\  C_i < \alpha\, C_{\mathrm{best}} \right\}$ where $\alpha\in(0,1)$ is a user-defined scalar that tightens the selection criterion by ensuring that an agent is chosen as the new best agent only if its cost is sufficiently lower than $C_{\mathrm{best}}$.

\subsection{GPU-accelerated Perception}

\begin{figure}[h]
    % \centering
    \includegraphics[width=\columnwidth]{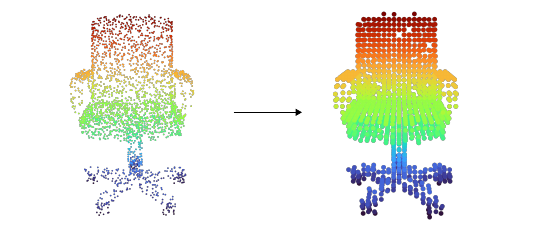}
    \caption{The perception pipeline transforms the merged point cloud $P_W$ (left) into the sphere-world approximation $\SWR$ (right) that enables robust Reactive Planning.}
    \label{fig:pc_to_sphere_world_representation}
\end{figure}

We propose a pipeline that creates a \emph{sphere-world approximation} \cite{rimon1992navigation} of the environment (Fig. \ref{fig:pc_to_sphere_world_representation}). We create a uniform lattice for our spherical obstacle primitives using voxel grid construction. Alternatively, it is possible to use non-uniform lattices, or non-spherical primitives \cite{gong2025geopf, eckhoff2023perception} for more control-compatible environment representations, but we defer that to future work. Our pipeline minimizes occlusions by fusing multi-view $3$D-sensing, to prevent unwanted reactive planner behavior such as piercing unobserved convex geometries. The volumetric discretization of the task-space geometry also offers uniformity for repulsive forces, resulting in smooth force interactions with obstacle primitives, unlike with raw point clouds where local point cloud density would distort motion. Our world model is not only suitable for trivial convex static scenarios, but (as we will show in Sec. \ref{sec:real_world_exp}) it also works in the real-world for non-convex, dynamic scenarios where reactive planners can often struggle, and where safe physical Human Robot Interaction (pHRI) is important.

Achieving high-fidelity perception at sensor frame rates is fundamentally a bandwidth and parallelism bound problem. CPUs, while possessing some Single Instruction, Multiple Data (SIMD) capability and parallelism, are not suited for processing large frames of point cloud data in real-time and risk starving other motion-planning critical threads. This is specially the case for a multi-agent reactive planning systems, like ours, that depends on CPU hyperthreads to manage agent threads and global state control. In contrast, GPUs are better-suited for processing perception tasks as they offer massive data-parallel execution for graphics workloads through a much larger number of simple processors. Therefore, we accelerate $3$D-data processing operations such as downsampling, bounding box filtering, voxel grid construction, and coordinate transforms using the GPU. This allows us to perform fast operations on high-density data. Incoming $3$D point cloud data is loaded onto the global memory (GPU DRAM) and remains resident there to help avoid frequent CPU memory reads and writes.

\subsection{SIMT-optimized Force Computation}\label{sec:vec_force_comp}

Existing approaches to calculating vector fields traditionally employ the CPU \cite{becker2021circular, becker2023informed, laha2021reactive} and are limited to a relatively small number of obstacles (on the order of tens to hundreds) to maintain the method's reactiveness. As the complexity of the scene grows, the need for a faster method for computing vector field forces also follows. Our simple world model, $\SWR$ offers several natural computational advantages for computing $\fr{}{k}$ for the heuristics in Sec. \ref{sec:vector_field_heuristics}. 

In our approach (Alg. \ref{alg:parallel_force}), force calculations are massively parallelized per obstacle by computing the force contribution of each obstacle on the agent in parallel, like in Fig. \ref{fig:complete_system_overview}. The key to our approach is to assign a single obstacle for processing from the global set $\SWR$ per GPU thread (line $5$). The shortest Euclidean distance is computed between the surface of the obstacle and the agent's current position (line $6$), and if the distance is smaller than $r_{d_i}$, a vector that represents the interaction between the agent and the obstacle is computed based on the associated vector equation (line $9$; see Sec. \ref{sec:vector_field_heuristics}). The partial result is written to shared memory (line $10$), and intra-block reductions are then performed, followed by an atomic accumulation into global memory to produce the net repulsive force $\fr{i}{k}$, which is returned as the response to the agent’s query (lines $15$-$21$). 
\newlength{\alglabelinputwidth}
\settowidth{\alglabelinputwidth}{\textbf{Input:}\ }

\newlength{\alglabeloutputwidth}
\settowidth{\alglabeloutputwidth}{\textbf{Output:}\ } 

\algrenewcommand{\algorithmiccomment}[1]{{\color{gray}\small // #1}}

\begin{algorithm}[]
\caption{\textsc{GPUForceCalculation} Kernel}
\label{alg:parallel_force}
\begin{algorithmic}[1]

\Statex \makebox[\alglabelinputwidth][l]{\textbf{Input:}}%
\parbox[t]{\dimexpr\linewidth-\alglabelinputwidth\relax}{%
End-effector position $\bm{p}$ and velocity $\dot{\bm{p}}$, the current goal position $\bm{p}_{\text{g}}$, $\SWR$ array, number of obstacles $\No$, radius of spherical primitive $r_{O}$, radius for the agent $r_{a_i}$, obstacle detection radius $r_{d_i}$, force scaling constant $k_f$, minimum acceptable distance $\epsilon$, a shared memory array $\bm{y}$, and thread block size $B$}

\Statex \makebox[\alglabeloutputwidth][l]{\textbf{Output:}}
\parbox[t]{\dimexpr\linewidth-\alglabeloutputwidth\relax}{%
A repulsive force vector $\fr{i}{k}$ at planning step $k$}

\State Initialize $b$ as the block index, and $t$ the thread index
\State $\bm{y}[t] \gets 0$ \Comment{Sets shared mem = 0}
\State $x \gets b\cdot B+t$ \Comment{Global obstacle index in $\mathcal{S}_W$}
\If{$x < {N}_{O}$}
    \State  $\bm{p}_{O_x}\gets \SWR[x]$                
    \State $d_{O_x} \gets \text{argmax}\big(\|\bm{p}_{O_x} \| - (r_{a_i} + r_{O}), \ \epsilon  \big)$
    \State $\fr{x}{}\gets0$
    \If{$d_{O_x}\le r_{d_i}$} \Comment{Calculate heuristics with Eq. \ref{eq:apf}, \ref{eq:mfi}}            
        \State $\fr{x}{} \gets$ \textsc{CALC\_HEURISTIC}$(\bm{p}, \dot{\bm{p}}, \bm{p}_g, \bm{p}_O)$
        \State $\fr{x}{}\gets k_f \cdot \fr{x}{}$
    \EndIf
    
    \State $\bm{y}[t] \gets \fr{x}{}$ \Comment{Store partial result in shared memory}
    
    \For{$z \gets \text{B}/{2}$ \textbf{downto} $1$ \textbf{by} $z \gets z/{2}$}
        \If{$t < z$ \textbf{and} $t + z < N_{O}$}
            \State $\bm{y}[t] \gets \bm{y}[t] + \bm{y}[t+z]$
        \EndIf
    \EndFor
    
    \If{$t = 0$}
      \State \textsc{ATOMIC\_ADD}$(\fr{i}{k}, \bm{y}[t])$
    \EndIf

\EndIf
\State \Return
\end{algorithmic}
\end{algorithm}

Each thread in our algorithm remains computationally lightweight because it performs only a small number of floating-point operations. This makes the kernel less likely to be compute bound than memory-latency bound. High occupancy due to high-resolution environment parallelism means that more warps (groups of threads) can run and cover memory delays (to ``hide'' warp-level latency). Our approach also stages results in shared memory and performs reductions while other warps are free to fetch obstacle data. Therefore, memory access and computation are overlapped at multiple levels, keeping more Streaming Multiprocessor (SM) units active. %From the perspective of time complexity, the parallelization scheme affords a near constant time complexity of approximately $O(1)$ as the compute time cost does not scale with scene complexity, satiating the time budget requirements for Problem P$2$\footnote{Note: A special CUDA kernel to calculate the Obstacle Distance Cost (Eq. \ref{eq:cost_obs}) leverages GPU-parallelism similar to Alg. \ref{alg:parallel_force} to calculate the SDF.}.

\textbf{Complexity Analysis.} Considering the number of obstacles in the environment $\No$, we assume the force calculations are executed on $P$ processor units. The kernel's time complexity per query can be estimated as $\bigO{\No/P}$. The atomic accumulation forces serial execution, and so the total time complexity becomes $\bigO{\No/P+\No} = \bigO{\No}$. The space complexity is $\bigO{\No}$, as our parallelization scheme does not affect the number of calculations.

\subsection{GPU-enabled Multi-Agent Compute Concurrency}

An ideal multi-agent planner should process forces for all agents efficiently, preferably in a fast, batched manner to maintain the requirement for active, parallel state exploration within a time budget (Problem P$3$). This batching can be implemented either as a single ``bulk'' kernel that handles all agents at once, or as multiple parallel specialized kernels for different agent heuristic types.

A bulk CUDA kernel approach would need to handle both environment and agent-level parallelism. For heterogeneous agents, a naive bulk kernel implementation is likely to incur warp divergence when threads in the same warp evaluate different heuristics. Complex heuristics (like those described in \cite{laha2023pmaf}) can involve a wide variety of per-agent dependencies and interactions. This increases kernel code size, and data layout and memory requirements. Larger kernels can lead to higher pressure on the SM's instruction cache and registers. This may reduce occupancy by increasing the time spent fetching data from global memory due to cache misses or register spilling, and limit the ability to hide memory latency. 
% To prevent a hit to performance, careful design—such as grouping agents by heuristic types and using CTAs—would be needed to mitigate warp divergence and maintain high throughput.

Our approach is simpler, leveraging specialized kernels for each heuristic type to concurrently process heterogeneous agents using parallel CUDA streams. Homogeneity across threads minimizes divergence within the warp. A smaller code footprint also reduces instruction cache misses and register pressure, potentially improving SM occupancy and latency hiding. Organizing the task-space agents by heuristic allows for efficient batching of force queries, and overlapped kernel execution and memory transfers for different heuristic types can help improve throughput. Our approach leverages code simplicity for memory efficiency and GPU throughput, particularly for diverse and complex agent behaviors. Although the CPU and GPU implementations employ identical planning algorithms, cost functions, and decision logic, the GPU implementation achieves higher success rates in some scenarios (e.g., the Trap scene). This difference arises from computational effects rather than algorithmic changes: the GPU enables faster planning cycles and more frequent cost-function evaluations, allowing agents to switch strategies and respond to emerging constraints in a more timely manner. In highly constrained or dynamic environments, this increased update frequency improves responsiveness and reduces failures due to delayed re-planning.

% \subsection{GPU-parallelized Distance Cost Computation} 
% The obstacle distance cost represents the agent’s proximity to surrounding obstacles. The calculation for this cost is offloaded to the Fields Computer Node to leverage GPU-parallelism similar to Alg. \ref{alg:parallel_force}'s. Sec. ~\ref{obs_dist_cost} provides the Signed Distance Field (SDF) equation used for the Obstacle Distance Cost kernel.
%%%%%%%%%%%%%%%%%%%%%%%%%%%%%%
% GEOMETRIC PLANNER
%%%%%%%%%%%%%%%%%%%%%%%%%%%%%%

%%%%%%%%%%%%%%%%%%%%%%%%%%%%%%
\section{Implementation}
%%%%%%%%%%%%%%%%%%%%%%%%%%%%%%

\subsection{Perception Pipeline}

We used \texttt{Open3D CUDA}\cite{zhou2018open3d} for GPU-accelerated rapid $3$D data processing operations. In the \emph{integration} stage, the pipeline ensures temporal alignment between $\Ns$ point cloud frames using a lossy latency-sensitive buffer that drops expired frames with a timeout threshold at approximately the sensor streaming rates. The pipeline downsamples, merges, and transforms point clouds. A merge of $\Ns$ unique views removes occlusions and obtains the unified point cloud $\Pw$. Filtering is done to retain points within the user-defined workspace $\mathcal{T}$. In the \emph{modeling} stage, FK is performed during runtime on a set of bounding spheres using live joint state data from the robot. We use the \texttt{foam} tool \cite{coumar2025foam} to create the sphere set offline for the robot morphology. An FK pass subtracts the points within the bounding sphere set using nearest neighbor search (NNS). The resulting point cloud, representing a $3$D bird's eye view (BEV) of the observed environment in the task-space (like in Fig. \ref{fig:complete_system_overview}), is voxelized with the voxel size $2r_{O}$. We filter the voxels to minimize spatio-temporal artifacts related to sensing. The voxel-center point clouds are extracted to yield a structured and volumetrically consistent $\SWR$.

\subsection{Planner Architecture}
The planner was implemented in C++ using the \texttt{gafro} library~\cite{low2023geometric} to implement the GA computations. It is worth highlighting that our framework and examples define the goal as an attractor directly on the rigid-body pose in $\mathrm{Spin}(3) \ltimes \mathbb{R}^3$ (double cover of SE($3$)). This explicitly generates attractor wrenches in se^*($3$), which are represented through the GA formulation---with pose error computed via logarithmic maps. This lets us treat translation and orientation within a single coupled operational-space objective, instead of handling them as separate control targets that must later be reconciled. We employ a dual-threaded architecture for the planner to control the robot body and compute the multi-agent trajectory rollout in parallel. The planner executes a \emph{planning loop}, wherein each iteration constitutes a planning cycle composed of $\Nk$ 
 steps. In each step, $\Na$ threads are spawned to ensure synchronous planning, lockstep progression, and uniform CPU allocation across all agents. As shown in Fig. \ref{fig:complete_system_overview}, the control thread is polled for the robot's current state at the start of every cycle. The agents within the cycle query the repulsive force kernels to get $\fr{i}{k}$. The control thread, alternatively, operates at approximately $100$ Hz, regularly retrieving $i^{*}$ from the planner and the physical robot state from low-level robot interfaces. At each control step, it queries the GPU Node with a copy of the current $i^{*}$'s configurations to get  $\fr{}{}$ in order to compute the end effector force for real-time robot control (Fig. \ref{fig:complete_system_overview}-right).

% %%%%%%%%%%%%%%%%%%%%%%%%%%%%%%
% % SIMULATIONS 
% %%%%%%%%%%%%%%%%%%%%%%%%%%%%%%

%%%%%%%%%%%%%%%%%%%%%%%%%%%%%%
\section{Simulation Experiments} \label{sec:sim_exp}
%%%%%%%%%%%%%%%%%%%%%%%%%%%%%%

In this section, we evaluate the performance of our reactive motion planner using synthetic simulations. We run our experiments on a PC running an Intel i$7$-$8565$U~$4$~cores~($8$~threads)~@~$1.8$~GHz Processor with an NVIDIA~RTX~$3070$~$8$~GB Graphics Card. As usual in the robotics literature~\cite{orthey2023sampling}, we rely on controlled, benchmarking-feasible simulated environments for quantitative evaluation. At the same time, challenging real-world experiments are performed to validate the end-to-end deployment and robustness.

\begin{table*}[t]
    \vspace{2mm}
    \caption{Performance comparison of GPU vs CPU implementations across different simulation scenarios. $\Na$, $\Nk$, $T_C$, and $T_M$ denote the number of planning agents, the number of planning steps, the computation time per step, and the motion duration, respectively.} 
    % \todo{add the table variables here: Na, Nc, Tc, TM.}
    %\todo{$T_C$ is already in the figure, same information twice, figure is much easier to look at, maybe convert the rest of the table to a figure as well?}} % I didn't convert into table because the table helps differentiate the impact of agents on Tc, look at how adding 1 more agent adds 20ms per 50 steps latency to the Tc for GPU cases. This is something hard to visualize, also I talked about in the text. 
    \label{tab:cpu_vs_gpu}
	\centering
	\resizebox{\textwidth}{!}{%
        \begin{tabular}{rrr|rrr|rrr|rrr}
			\hline
			\textbf{Scene} & 
			$\Na$&
            \textbf{Device} &
            \multicolumn{3}{c|}{$\Nk=50$} &
			\multicolumn{3}{c|}{$\Nk=100$} &
			\multicolumn{3}{c}{$\Nk=200$} \\
			&  &  &
			Succ. & $T_{C}$ [ms] & $T_{M}$ [s] &
			Succ. & $T_{C}$ [ms] & $T_{M}$ [s] &
			Succ. & $T_{C}$ [ms] & $T_{M}$ [s] \\
			\hline
			\multirow{2}{*}{Trap}      & \multirow{2}{*}{2}  &   GPU & \textbf{30\%}  & \textbf{119.76} $\pm$ \textbf{9.12} & 112.64 $\pm$ 0.00  & \textbf{50\%}  & \textbf{238.89} $\pm$ \textbf{19.97} & 112.64 $\pm$ 0.01  & \textbf{70\%}  & \textbf{478.55} $\pm$ \textbf{36.08} & 112.14 $\pm$ 0.01  \\
			                           &                        &   CPU & \textbf{30\%}  & 206.98 $\pm$ 16.56                  & 112.63 $\pm$ 0.00  & 40\%           & 420.14 $\pm$ 22.22                   & 112.25 $\pm$ 0.22  & 40\%           & 836.72 $\pm$ 42.98                   & 111.62 $\pm$ 0.00  \\\hline
			\multirow{2}{*}{Opening}       & \multirow{2}{*}{1}  & GPU & \textbf{100\%} & \textbf{101.05} $\pm$ \textbf{2.54} & 104.68 $\pm$ 3.79  & \textbf{100\%} & \textbf{201.57} $\pm$ \textbf{3.62}  & 104.78 $\pm$ 3.85  & \textbf{100\%} & \textbf{403.42} $\pm$ \textbf{6.42}  & 104.83 $\pm$ 4.03  \\
			                           &                        & CPU & 90\%           & 249.8 $\pm$ 14.06                   & 105.01 $\pm$ 5.02  & 90\%           & 499.45 $\pm$ 25.04                   & 103.39 $\pm$ 4.63  & 90\%           & 999.86 $\pm$ 42.15                   & 103.67 $\pm$ 4.55  \\\hline
			\multirow{2}{*}{Cluttered} & \multirow{2}{*}{1}  & GPU & \textbf{100\%} & \textbf{100.96} $\pm$ \textbf{2.57} & 107.48 $\pm$ 13.05 & \textbf{100\%} & \textbf{203.04} $\pm$ \textbf{4.28}  & 107.38 $\pm$ 13.06 & \textbf{100\%} & \textbf{405.55} $\pm$ \textbf{7.28}  & 106.63 $\pm$ 12.92 \\
			                           &                        & CPU & \textbf{100\%} & 401.98 $\pm$ 21.81                  & 106.12 $\pm$ 12.98 & \textbf{100\%} & 802.86 $\pm$ 37.24                   & 105.82 $\pm$ 12.83 & \textbf{100\%} & 1,599.15 $\pm$ 70.03                 & 105.77 $\pm$ 12.51 \\\hline
			\multirow{2}{*}{Passage}   & \multirow{2}{*}{2}  & GPU & \textbf{100\%} & \textbf{120.49} $\pm$ \textbf{9.34} & 108.33 $\pm$ 0.24  & \textbf{100\%} & \textbf{241.29} $\pm$ \textbf{16.81} & 108.28 $\pm$ 0.23  & \textbf{100\%} & \textbf{481.96} $\pm$ \textbf{30.48} & 107.73 $\pm$ 0.20  \\
			                           &                        & CPU & 70\%           & 676.10 $\pm$ 41.53                  & 107.69 $\pm$ 0.17  & 60\%           & 1,345.46 $\pm$ 60.26                 & 107.37 $\pm$ 0.25  & 70\%           & 2,680.91 $\pm$ 101.83                & 105.76 $\pm$ 0.23  \\\hline
		\end{tabular}
	}
\end{table*}

For a thorough evaluation of our system, we look at quantitative measurements of 1) the runtime performance of the planner in GPU and CPU mode, and 2) the success rate of the multi-agent planner versus few-agents scenarios.%  scheme compared to the aforementioned single-agent methods.  
% We use single-agent heuristics outlined in Eq. \ref{eq:apf} (APF) and Eq. \ref{eq:mfi} (MFI) as base cases. We also create multi-agents that use a composition of these heuristics, which we refer to as MULTI in the tables.
% We use c

\begin{figure}[]
    \centering
    \includegraphics[width=\columnwidth]{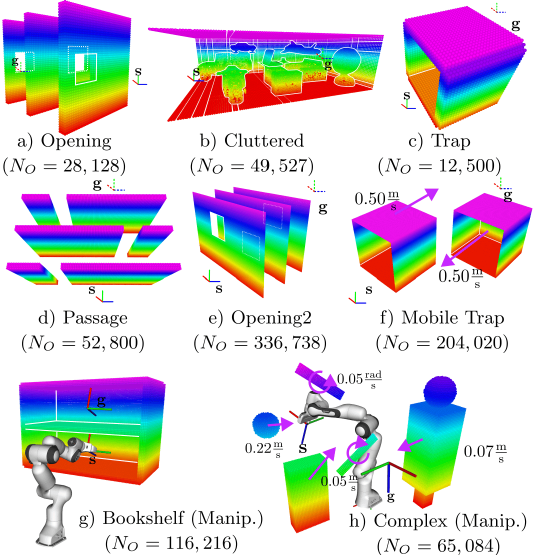}
    \caption{Environments used for simulation. These include: a) Opening, b) Cluttered,  c) Trap, and d) Passage}
    \label{fig:simulation_scenes}
\end{figure}

Fig. \ref{fig:simulation_scenes} shows the simulation scenes used in this evaluation. %, with Table II containing the scene descriptions. 
These scenes are artificially generated and high-fidelity (many obstacles) to resemble the types of environments the robot may encounter in the real world. These environments also have characteristics that affect goal convergence for planners, such as concavity, narrow passages, and dynamic objects. %To study the planning algorithm in isolation, we set the simulation scenes to be static environments. Dynamic environments are considered qualitatively in Section \ref{sec:real_world_exp}.

% --- CPU vs GPU ---
\subsection{CPU vs GPU Computation Performance Comparison}

\begin{figure}[htbp]
    % \vspace{-0.25cm}
    \includegraphics[width=\columnwidth]{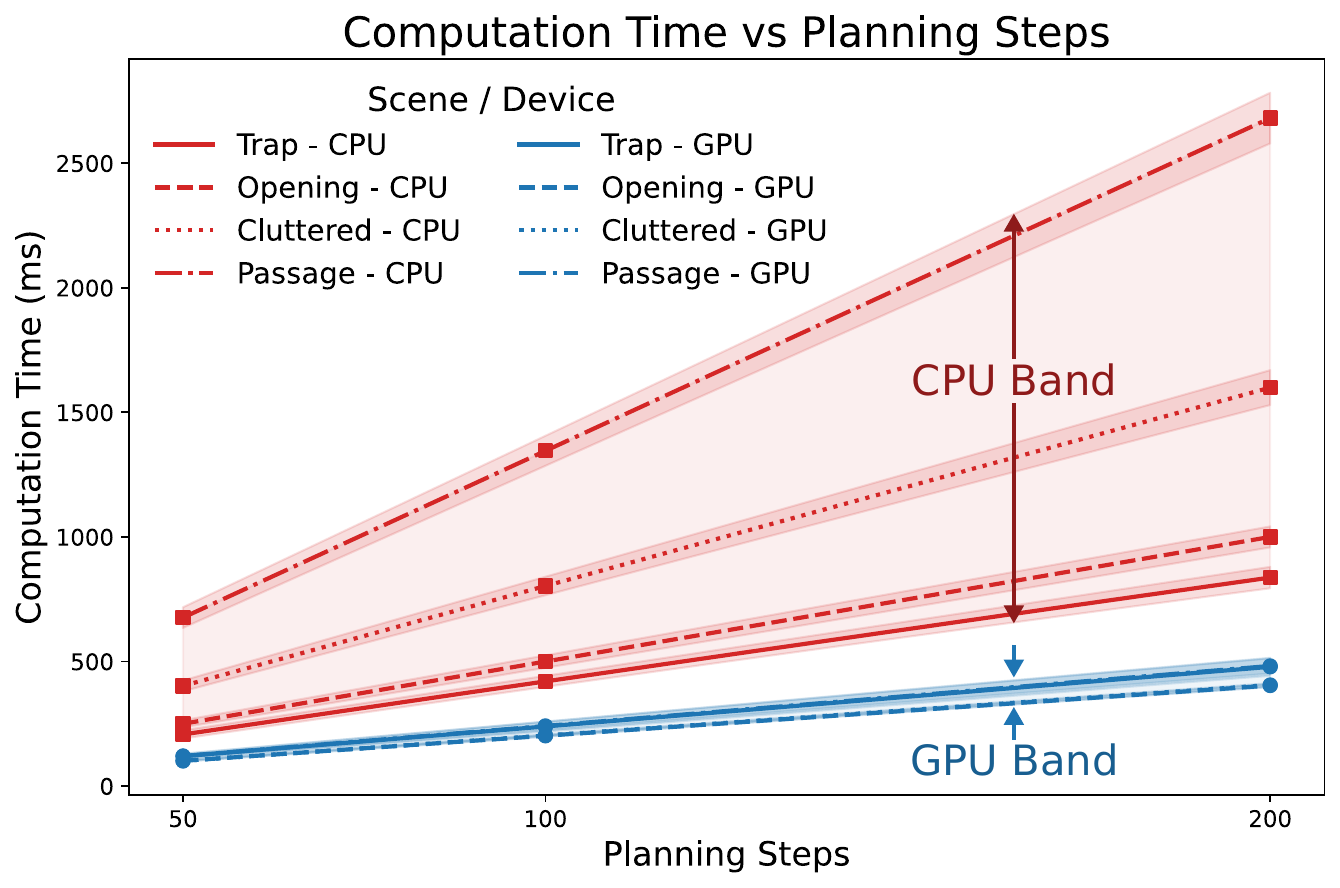}
    \caption{Computation time (mean ± SD) for planning steps (50, 100, 200) across four scenes (Trap, Opening, Cluttered, and Passage) on GPU (blue) and CPU (red). The slimmer GPU band demonstrates superior scaling versus the CPU w.r.t. scene fidelity and agent count.
    }
    \label{fig:cpu_vs_gpu}
\end{figure}

In Table \ref{tab:cpu_vs_gpu}, we measure the computation time ($T_{C}$), motion duration ($T_M$), and success rate in GPU and CPU modes for varying planning steps $\Nk\in\{50,100,200\}$. We maintain the same agent configurations across GPU and CPU trials. Each case in the table represents $10$ trials. The $T_C$ includes the communication overhead from ROS $2$.

We note that the $T_{C}$ on the GPU is smaller compared to the CPU by at least a factor of $2$. We also note that the trends for $T_{C}$ are largely invariant to the fidelity of the scene. The GPU version of the planner has an expected performance guarantee of approximately $T_{C}/\Nk = 2$ ms for each planning step. The standard deviations for the GPU version are also smaller than those of the CPU and nearly zero in all cases. We note that the addition of multi-agents ($\Na$) incurs a penalty of approximately $0.4$ ms per planning step, while it nearly doubles the $T_{C}$ of the CPU version. Fig.~\ref{fig:cpu_vs_gpu} illustrates the increase in $T_{C}$ and the spread of $T_{C}$ for CPU and GPU with respect to $N_{k}$.

% We also note that the computational efficiency of the GPU version (in generating plans faster than the CPU through parallelized force calculations) also increases the success rate of the planner, such as in the trap scenario. Where the longer planning cycles increase the success rate for the GPU version. We suspect this to be the case because the GPU version (compared to its CPU counterpart) is able to finish the planning cycle faster and allows the cost function to decide when to switch agents.

The GPU version’s improved computational efficiency—due to faster, parallelized force calculations—also boosts the planner's success rate, as seen in the trap scenario, where longer planning cycles benefit the GPU version. We suspect this is because the GPU version completes the planning cycle faster than the CPU version, allowing the cost function to determine when to switch agents.

% --- Multiagent Performance --- 
\subsection{Quantitative Evaluation for Multi-Agents}

In Table II, we measure motion duration $T_{M}$, computation time $T_{C}$, agent switches $\Nsw$, minimum obstacle distance $L_{\mathrm{min}}$, and success rate for on the GPU. %single-agent and multi-agent schemes on the GPU.  
We compare the performance of twin-agent heuristics with multi-agent (ensemble) strategies (MULTI) containing heuristic constructions from~\cite{laha2023pmaf}. We optimize the APF and MFI methods for goal-convergence success and use the same configurations for the MULTI method, for which we optimize the cost function parameters for $\Nsw$ and success rate. Each case in the table represents $15$ trials. Across all static and dynamic scenes, multi-agent planners (MFI and MULTI) consistently outperform the single-method APF baseline in success rate and solution quality; APF fails in complex scenarios despite low compute cost. Both MFI and MULTI perform robustly, with MULTI showing the best scalability in challenging settings. In Opening2, MULTI improves success over MFI ($73.3$ vs. $66.7$), highlighting the benefits of heterogeneous agents. In dynamic scenes (Mobile Trap and Complex), MULTI matches or exceeds MFI’s high success rates while maintaining comparable planning times. In the most constrained scenarios, MULTI achieves higher success rates and fewer obstacle-clearance violations, demonstrating greater robustness. Although computationally more expensive, MULTI’s overhead yields substantially higher reliability in difficult environments (closer to the real-world).

\begin{table}[htbp]
% \cob{\caption{Detailed quantitative analysis of different planners across test scenarios. MFI and MULTI generally achieve better cost metrics compared to APF, particularly in complex scenarios like Passage where APF fails completely.}}

\caption{Quantitative evaluation of planning schemes with differing multi-agent initializations across test scenarios. $\Nk$, $\Na$, $T_C$, $T_M$, and $L_{\mathrm{min}}$ denote the number of planning steps, number of planning agents, the computation time per step, and the motion duration, respectively.}

\label{tab:planner_comp}
\resizebox{\columnwidth}{!}{
    
    \begin{tabular}{ccccccccc}
    \hline
    \multicolumn{1}{l}{\textbf{Scene}} & \multicolumn{1}{l}{\textbf{Method}} & \multicolumn{1}{l}{\bm{$\Nk$}} & \multicolumn{1}{l}{\bm{$\Na$}} & \multicolumn{1}{l}{\bm{$T_{C} [ms]$}} & \multicolumn{1}{l}{\bm{$T_{M} [s]$}} & \multicolumn{1}{l}{\bm{$L_{min} [m]$}} & \multicolumn{1}{l}{\bm{$\Nsw$}} & \multicolumn{1}{l}{\textbf{Succ.}} \\ \hline
    \multirow{3}{*}{\shortstack[l]{Opening2 \\ (Static)}} & APF & 100 & 2 & $145.33 \pm 0.00$ & $0.00 \pm 0.00$ & \bm{$0.75 \pm 0.01$} & $0.00 \pm 0.00$ & $0.00$\% \\
     & MFI & 100 & 2 & $184.43 \pm 0.00$ & \bm{$23.27 \pm 2.21$} & $0.24 \pm 0.03$ & $2.20 \pm 3.21$ & $66.67$\% \\
     & MULTI & 100 & 11 & $257.01 \pm 0.00$ & $24.77 \pm 1.59$ & $0.24 \pm 0.03$ & $7.40 \pm 1.74$ & \bm{$73.33$\%} \\ \hline
    \multirow{3}{*}{\shortstack[l]{Mobile Trap \\ (Dynamic)}} & APF & 150 & 2 & $156.97 \pm 28.81$ & $0.00 \pm 0.00$ & \bm{$0.33 \pm 0.09$} & $0.53 \pm 1.02$ & $0.00$\% \\
     & MFI & 150 & 2 & $178.64 \pm 24.48$ & \bm{$21.78 \pm 1.74$} & $0.26 \pm 0.05$ & $2.73 \pm 1.84$ & \bm{$100.00$\%} \\
     & MULTI & 150 & 11 & $227.29 \pm 43.41$ & $22.68 \pm 1.76$ & $0.22 \pm 0.06$ & $11.07 \pm 2.17$ & \bm{$100.00$\%} \\ \hline
    \multirow{3}{*}{\shortstack[l]{Bookshelf \\ (Static)}} & APF & 200 & 2 & $230.85 \pm 18.64$ & $0.00 \pm 0.00$ & \bm{$0.11 \pm 0.00$} & $13.27 \pm 0.57$ & $0.00$\% \\
     & MFI & 200 & 2 & $250.48 \pm 17.51$ & $16.53 \pm 0.32$ & $0.07 \pm 0.00$ & $6.13 \pm 0.72$ & \bm{$100.00$\%} \\
     & MULTI & 200 & 7 & $469.82 \pm 23.82$ & \bm{$10.99 \pm 0.57$} & \bm{$0.11 \pm 0.00$} & $12.13 \pm 1.26$ & \bm{$100.00$\%} \\ \hline
    \multirow{3}{*}{\shortstack[l]{Complex \\ (Dynamic)}} & APF & 75 & 2 & $97.10 \pm 17.26$ & $0.00 \pm 0.00$ & \bm{$0.13 \pm 0.01$} & $0.00 \pm 0.00$ & $0.00$\% \\
     & MFI & 75 & 2 & $100.70 \pm 18.09$ & $9.59 \pm 3.04$ & $0.07 \pm 0.04$ & $13.07 \pm 8.86$ & $80.00$\% \\
     & MULTI & 75 & 10 & $173.03 \pm 43.68$ & \bm{$8.20 \pm 1.80$} & $0.07 \pm 0.04$ & $25.87 \pm 13.00$ & \bm{$93.33$\%} \\ \hline
    \end{tabular}
    
}
\end{table}

We note that the goal convergence success rate of the multi-agent scheme is almost always on par with the best single agent, and attribute the $\Nsw$ to be the determining contributor. We note that the $T_{M}$ of the multi-agent method often falls in between the $T_M$ of the single-agent methods, with the cluttered scene being an example of where it outperforms both agents. We note that the small $\Nsw$ for the passage scenario correlates with the slightly lower success rate for that case. We also note that the $T_{C}$ for all cases in the table is consistent with the $T_{C}$ from the GPU cases in Table \ref{tab:cpu_vs_gpu}.

\section{Real-world Experiments}  
\label{sec:real_world_exp}
In this section, we present real-world experimental evaluation of the planner and perception framework. We do not focus on whole-body avoidance, but it can be integrated in the system, as shown in \cite{becker2023informed}. We run the experiments on an Ubuntu 24.04 PC running with an Intel i$7$-$9800$X $8$ cores ($16$~threads)~@~$3.80$~GHz Processor with an Nvidia RTX~$2080$Ti $11$~GB Graphics Card with ROS Rolling; connected to a Franka Emika Panda $7$-DOF manipulator robot and two Intel RealSense stereo depth cameras streaming $848\times480$ resolution depth frames at $60$~FPS.
\begin{figure}[htbp]    
    \centering
    \includegraphics[width=\columnwidth]{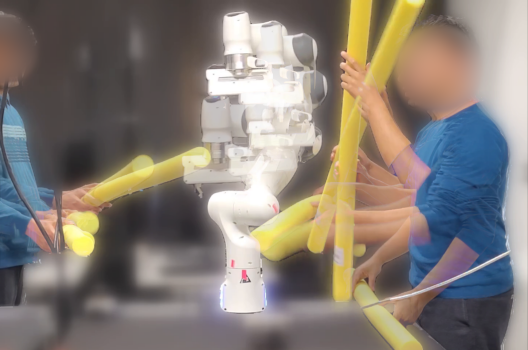}
    \caption{Our framework successfully achieves fast reactive motion on $7$-DOF Franka Panda setup, enabling dynamic obstacle avoidance. Shown here is a superimposed image of $2.5$ seconds of motion where two human actors actively perturb the arm's motion with pool noodles.}
    \label{fig:rwe_reactive_motion}
\end{figure}
We conduct two qualitative experiments to evaluate the robot's performance. The first experiment involves a concave trap that the robot must avoid while navigating toward a goal on the opposite side (Fig.~\ref{fig:front_page_graphic}). This scenario presents a significant challenge for reactive planners due to the local minima problem, testing the robot's ability to navigate effectively in complex environments. The second experiment features two humans who actively provide disturbances, testing the robot's ability to evade these interruptions while maintaining safe and reactive behavior (Fig.~\ref{fig:rwe_reactive_motion}). In both cases, the robot successfully adapts to the dynamic conditions, demonstrating its robustness in real-world applications. A video of these experiments can be viewed in the attached supplement.

%%%%%%%%%%%%%%%%%%%%%%%%%%%%%%
% CONCLUSION
%%%%%%%%%%%%%%%%%%%%%%%%%%%%%%
%%%%%%%%%%%%%%%%%%%%%%%%%%%%%%
\section{Conclusion, Limitations, And Future Work}
%%%%%%%%%%%%%%%%%%%%%%%%%%%%%%
In this letter, we present a perception-planning framework for real-time reactive motion generation. The proposed GPU-accelerated pipeline aggregates and transforms multiple temporally synchronized point cloud frames into a structured vector representation, which is then leveraged by a task-space reactive planner to generate adaptive motion behaviors. While geometric algebra (GA) provides a concise formulation for representing the $\mathrm{Spin}(3) \ltimes \mathbb{R}^3$ attractor and the associated coupled pose error, the primary emphasis of this work is on the proposed GPU-based perception-to-planning integration, which enables low-latency, closed-loop reactivity in dynamic environments. Accordingly, the experimental evaluation focuses on demonstrating reactivity in these challenging scenarios, rather than providing a comprehensive treatment of higher-dimensional repulsive terms. A full exploration would require additional tooling, visualization, and analysis to be meaningfully interpreted along executed robot trajectories. The resulting motions demonstrate robustness to reactive disturbances and environmental uncertainties, and local minima traps. At present, the approach incurs nontrivial CPU and ROS-related overhead, which could be significantly reduced by implementing tensorized motion planning. Future work will also investigate whether the proposed representation supports task-dependent approximation fidelity, the use of policy networks for automatic agent tuning, active inference for improved environment perception, and generalization across diverse robot morphologies.

\bibliographystyle{IEEEtran}
\bibliography{bibliography}

\end{document}